\documentclass[runningheads,a4paper]{llncs}

\usepackage{graphicx}
\usepackage{amsmath}
\usepackage{amssymb}
\usepackage[compatibility=false]{caption}
\usepackage{subcaption}

\usepackage{float}

\usepackage{enumerate}
\usepackage{floatflt}

\usepackage{url}
\urldef{\mailsa}\path|{giuseppe.airof, rosaria.francesca.rossini}@gmail.com|
\urldef{\mailsb}\path|{tiziana.armano, anna.capietto, nadir.murru}@unito.it|    
\newcommand{\keywords}[1]{\par\addvspace\baselineskip
\noindent\keywordname\enspace\ignorespaces#1}

\begin{document}

\mainmatter

\title{Artificial neural networks and fuzzy logic for recognizing alphabet characters and mathematical symbols}

\titlerunning{Neural networks and fuzzy logic for recognizing characters}

\author{Giuseppe Air\`o Farulla$^1$\and Tiziana Armano$^2$\and Anna Capietto$^2$\and Nadir Murru$^2$\and Rosaria Rossini$^3$
\thanks{This work has been developed in the framework of an agreement between IRIFOR/UICI (Institute for Research, Education and Rehabilitation/Italian Union for the Blind and Partially Sighted) and Turin University.}}
\authorrunning{Air\`o Farulla, Armano, Capietto, Rossini, Murru}
\institute{$^1$Department of Control and Computer Engineering, Politecnico di Torino,\\
Corso Duca degli Abruzzi 24, 10129, Torino, Italy\\
$^2$Department of Mathematics, University of Turin,\\
Via Carlo Alberto 10, 10121 Torino, Italy\\
$^3$Istituto Superiore Mario Boella, Center for Applied Research on ICT,\\
Via Pier Carlo Boggio 61, 10138, Torino, Italy\\
\mailsa\\
\mailsb\\}

\maketitle

\begin{abstract}
Optical Character Recognition software (OCR) are important tools for obtaining accessible texts. We propose the use of artificial neural networks (ANN) in order to develop pattern recognition algorithms capable of recognizing both normal texts and formulae. We present an original improvement of the backpropagation algorithm. Moreover, we describe a novel image segmentation algorithm that exploits fuzzy logic for separating touching characters.
\end{abstract}

\section{Introduction}
Currently, InftyReader is the unique OCR that performs automatic recognition of both normal texts and formulae \cite{Infty}. The pattern recognition algorithm used by the authors of InftyReader is based on support vector machine learning \cite{Infty}. In this paper, we propose to base the pattern recognition algorithm on artificial neural networks (ANNs). ANNs can be successfully exploited for developing pattern recognition algorithms \cite{Bis}. Recently, several studies have been pointed out on ANNs for the automatic recognition of characters in different alphabets, such as Latin \cite{Shri}, Arabic \cite{Sah} and many others. However, it appears that ANNs have not been exploited for recognizing both alphabet characters and mathematical symbols in printed documents yet, since the large amount of different patterns to recognize does not allow fast convergence of the training algorithms. In this paper, we address this problem by means of an original use of the Kalman filter (mainly based on the work of Murru and Rossini \cite{Mur}), with the aim of improving the rate of convergence in the training of ANNs even in presence of a large amount of patterns that must be recognized. In particular we study the backpropagation (BP) algorithm. It is well--known that convergence of BP is heavily affected by initial weights \cite{Adam}. Different initialization techniques have been proposed such as adaptive step size methods \cite{Sch}, partial least squares method \cite{Hsiao}, interval based method \cite{Sod}. Other different approaches can be found, e.g., in \cite{Erd}, \cite{Adam2}. However, random weight initialization is still the most used method also due to its simplicity. Thus, the study of new initialization methods is an important research field in order to improve application of neural nets and deepen their knowledge. 

Within an OCR, pattern segmentation is a required step before pattern analysis and recognition. The most common character segmentation algorithms are based on vertical projection, pitch estimation or character size, contour analysis, or segmentation--recognition coupled techniques \cite{Lu}. Several methods for separating touching characters have been developed, see, e.g., \cite{Liang}, \cite{Utpal}, \cite{Saba}, \cite{Kum}. In presence of formulae, separation of touching characters is harder since cutting positions can occur vertically, horizontally and diagonally. Several methods perform differently according to the features of the characters and their selection and use is an art rather than a technique. In other words, features selection mainly depends on the experience of the authors. Thus, in this context, fuzzy logic can be very useful, since it is widely used in applications where tuning of features is based on experience and it can be preferred to a deterministic approach. In this paper, we propose a method that combines, by means of a fuzzy logic based approach, some state--of--the--art features usually exploited one at a time.

In Section \ref{sec:bp}, we explain a novel method for the initialization of weights that improves performances of the backpropagation algorithm in order to train neural nets in the field of pattern recognition. In Section \ref{sec:fuzzy}, we present a fuzzy based approach for performig segmentation of touching characters. Finally, Section \ref{sec:results} and \ref{sec:conc} are devoted to numerical results and conclusions, respectively.

\section{An improvement of the backpropagation algorithm} \label{sec:bp}
Let us consider a neural network with $L$ layers. Let $N(i)$ be the number of neurons in the layer $i$, for $i=1,...,L$, and $w^{(k)}_{ij}$ be the weight of connection between the $i$--th neuron in the layer $k$ and the $j$--th neuron in the layer $k-1$.
An ANN is trained over a set of inputs so that it provides a fixed output for a given training input. Let us denote $X$ the set of training inputs. An element $\textbf x\in X$ is a vector (e.g., a string of bits representing a pattern).
Let $a_i^{(k,\textbf x)}$ be the output of the $i$--th neuron in layer $k$ when an input $\textbf x$ is processed by the ANN. This output is computed as follows:
$$\begin{cases} a_i^{(1,\textbf x)}=f(x_i) \cr  a_i^{(k,\textbf x)}=f\left(\sum_{j=1}^{N(k-1)}w_{ij}^{(k)}a_j^{(k-1,\textbf{x})}\right),\quad k=2,...,L, \end{cases}$$
where $f$ is the activation function (usually, the sigmoidal or hyperbolic tangent function). 
Finally, let $\textbf y^{(\textbf x)}$ be the desired output of the neural network corresponding to the input $\textbf x$. In other words, we would like that $\textbf a^{(L,\textbf x)}=\textbf y^{(\textbf x)}$, when neural net processes input $\textbf x$. Clearly, this depends on weights $w^{(k)}_{ij}$ and it is not possible to know their correct values a priori. Thus, it is usual to randomly initialize values of weights and use a training algorithm in order to adjust their values. One of the most common and used training algorithm is the BP. In \cite{Mur}, Murru and Rossini proposed an original approach for the initialization of weights mainly based on a customization of the Kalman filter through a Bayesian approach, in order to improve performances of BP algorithm.
The Kalman filter is a well--established technique to estimate the state $\textbf w_t$ of a dynamic process at each time $t$. Specifically, the Kalman gain matrix balances prior estimations and measurements so that an estimation is provided as follows: $\mathbf{\tilde w}_t=\textbf w^-_t+K_t(\textbf m_t-\textbf w^-_t)$, where $\textbf m_t$ is a measurement of the process, $\textbf w^-_t$ a prior estimation of the process, $K_t$ the Kalman gain matrix. 
As in \cite{Mur}, we customize the Kalman filter modeling measurements and prior estimations by means of multivariate normal random variables $\textbf W_t$ and $\textbf M_t$ such that their density functions satisfy
$g(\textbf W_t)=\mathcal N(\textbf w_t^-,Q_t),$ $g(\textbf M_t \vert \textbf W_t)=\mathcal N(\textbf m_t^-,R_t),$
where $Q_t$ and $R_t$ are covariance matrices conveniently initialized. In this way, the posterior density is given by
$g(\textbf W_t|\textbf M_t)\propto \mathcal N(\textbf w_t^-,Q_t)\mathcal N(\textbf m_t, R_t)=\mathcal N(\mathbf{\tilde w}_t,P_t),$
where
$ \mathbf{\tilde w}_t=(Q_t^{-1}+R_t^{-1})^{-1}(Q_t^{-1}\textbf w_t^-+R_t^{-1}\textbf m_t)$, $P_t=(Q_t^{-1}+R_t^{-1})^{-1}. $
We can apply this technique to weights initialization considering processes $\textbf w_t(k)$, for $k=2,...,L$, as non--time--varying quantities
whose components are the unknown values of weights $w^{(k)}$, for $k=2,...,L$, of the neural net such that $\textbf a^{(L,\textbf x)}=\textbf y^{(\textbf x)}$. The goal is to provide an estimation of initial weights to reduce the number of steps that allows convergence of BP neural net. Thus, for each set $w^{(k)}$ we consider initial weights as unknown processes and we optimize randomly generated weights (which we consider as measurements of the processes) with the above approach. 
In these terms, we derive an optimal initialization of weights by means of the following equations:
\begin{equation} \label{eq:alg}\begin{cases}\textbf m_t=Rnd(-h,h) \cr (R_t)_{ii}=\cfrac{1}{N(k)N(k-1)}\sum_{\textbf x\in X}\| \textbf{d}^{(k,\textbf x)} \|^2, \quad (R_t)_{lm}=0.7 \cr  \mathbf{\tilde w}_t=(Q_t^{-1}+R_t^{-1})^{-1}(Q_t^{-1}\textbf w_t^-+R_t^{-1}\textbf m_t) \cr Q_{t+1}=(Q_t^{-1}+R_t^{-1})^{-1},\quad \textbf w_{t+1}^-=\mathbf{\tilde w}_t \end{cases}\end{equation}
where $Rnd(-h,h)$ denotes the function that samples a random real number in the interval $(-h,h)$ and $ \textbf{d}^{(k,\textbf x)}$ is the usual error of the $k$--th layer when an input $\textbf x$ is given.

\section{Image segmentation with fuzzy logic}\label{sec:fuzzy}
In a binarized image, a pattern can be represented by a matrix whose entries are 0 (white pixels) and 1 (black
pixels). Generally, methods for segmenting touching characters define a function based on some features that characterize cut positions. Then, such a function is evaluated for each column (row, or diagonal) of the matrix and the cut position is chosen depending on its values. Classical functions of this kind are the peak--to--valley function $g$ and the function $h$ defined as
$$g(i)=\cfrac{V(l_i)-2V(i)+V(r_i)}{V(i)+1},\quad h(i)=\cfrac{V(i-1)-2V(i)+V(i+1)}{V(i)},$$
where $V(i)$ denotes the vertical projection function for the $i$-th column (row or diagonal), $l_i$ and $r_i$ are the peak positions on the left side and right side of $i$, respectively. Let us denote by $\bar g$ and $\bar h$ the functions $g$ and $h$ normalized to $[0,1]$, respectively. In the following, we will consider $\tilde g = 1-\bar g$ and $\tilde h = 1-\bar h$. Here, we propose a fuzzy routine typically identifying a column, a row, or a diagonal of the matrix that could be a cut point that conveniently separates the touching characters. 

Our fuzzy routine combines four state--of--the--art features: distance from the center of the pattern; crossing count, i.e., the number of transitions from white to black pixels, and viceversa; the function $\bar g$; the function $\bar h$. These features are widely used in literature for determining cut positions in touching characters \cite{Liang}. However, they are usually managed separately and in a deterministic way. In our approach, these features are combined by means of convenient fuzzy rules in order to exploit the information given by each of these features. 

Given a pattern in a binarized image, let $A$, $m$, $n$, and $c$ be the matrix of pixels of the binarized matrix, the number of rows of $A$, the number of columns of $A$, and the central column of $A$, respectively. For the sake of simplicity, in the following we only focus on columns of $A$ and when we refer to a column $i$ of $A$, we refer to the $i$--th column of $A$, i.e., we are considering the vector of length $m$ whose elements are the entries of the $i$--th column. For each column $i$ of $A$, we define its normalized distance from the center of the pattern as $d(i)=\frac{\lvert c-i \rvert}{c}$. Moreover, we indicate the crossing count function by $f$, i.e., the number of transitions between white and black pixels.

\begin{figure*}[t!]
    \centering
    \begin{subfigure}[t]{0.42\textwidth}
        \centering
        \includegraphics[width=\textwidth]{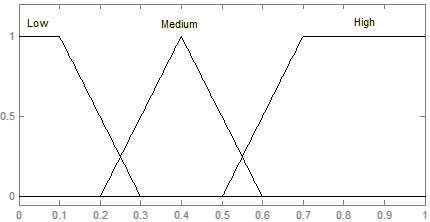}
        \caption{\label{fig:mf-d}}
    \end{subfigure}%
    ~ 
    \begin{subfigure}[t]{0.405\textwidth}
        \centering
        \includegraphics[width=\textwidth]{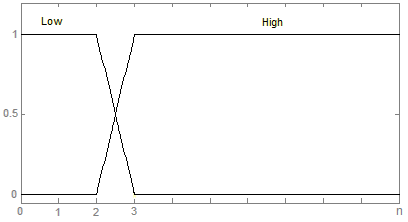}
        \caption{\label{fig:mf-f}}
    \end{subfigure}%
    ~ 
    \begin{subfigure}[t]{0.42\textwidth}
        \centering
        \includegraphics[width=\textwidth]{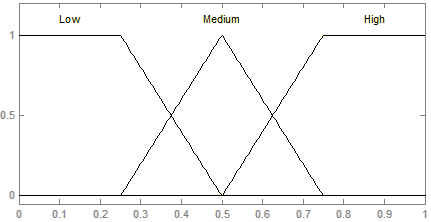}
        \caption{\label{fig:mf-g-h-o}}
    \end{subfigure}
    \caption{Membership functions of the fuzzy sets related to, respectively, (\subref{fig:mf-d}) $d$,  (\subref{fig:mf-f}) $f$, (\subref{fig:mf-g-h-o})  $\tilde g$, $\tilde h$, and $\rho$.}
\end{figure*}

To design a suitable fuzzy strategy, some steps are required, in order to introduce the notion of a fuzzy degree qualifying a column $i$ to be a cut position: for short $\rho = \rho(i) \in [0,1]$. In our model, the low values of $\rho$ locate good cut positions. The strategy can be detailed by means of the fuzzification of the functions $d$, $f$, $\tilde g$, $\tilde h$. Figures \ref{fig:mf-d}, \ref{fig:mf-f}, \ref{fig:mf-g-h-o} show the fuzzy sets and the related membership functions. Note that we have considered the same fuzzy sets and membership functions for $\tilde g$, $\tilde h$, and $\rho$. 

For each column $i$ of $A$, the inference system combines the values $d(i)$, $f(i)$, $\tilde g(i)$, $\tilde h(i)$ and produces the fuzzy output $\rho(i)$ by means of the following fuzzy rules:

\begin{enumerate} \small
\item if $d(i)$ is Low and $\tilde g(i), \tilde h(i)$ are not High and $f(i)$ is Low, then $\rho(i)$ is Low;
\item if $\tilde g(i), \tilde h(i)$ are Low and $d(i)$ is Medium and $f(i)$ is Low, then $\rho(i)$ is Low;
\item if $\tilde g(i)$ is Low and $d(i)$ is not High and $\tilde h(i)$ is not Low and $f(i)$ is Low, then $\rho(i)$ is Low;
\item if $d(i)$ is Low and $\tilde g(i), \tilde h(i)$ are not High and $f(i)$ is High, then $\rho(i)$ is Medium;
\item if $\tilde g(i), \tilde h(i)$ are Low and $d(i)$ is Medium and $f(i)$ is High, then $\rho(i)$ is Medium;
\item if $\tilde g(i)$ is Low and $d(i)$ is not High and $\tilde h(i)$ is not Low and $f(i)$ is High, then $\rho(i)$ is Medium;
\item if $\tilde h(i)$ is Low and $d(i)$ is not High and $g(i)$ is not Low and $f(i)$ is Low, then $\rho(i)$ is Medium;
\item if $d(i), \tilde g(i), \tilde h(i)$ are Medium and $f(i)$ is Low, then $\rho(i)$ is Medium;
\item otherwise $\rho(i)$ is High.
\end{enumerate}

These fuzzy rules have been tuned using heuristic criteria taking into account that high values of $g$, $h$ and low values of $d$, $f$ usually identify cut positions. The inference engine is the basic Mamdani model with if--then rules, minimax set--operations, sum for composition of activated rules and defuzzification based on the centroid method. The Mamdani model is congenial to capture and to code expert-based knowledge.

\section{Numerical results}\label{sec:results}

In this section, we describe the process to train neural networks in order to recognize both characters and mathematical symbols, comparing the rate of convergence of the BP algorithm with the Bayesian initialization (BI) presented in Section \ref{sec:bp} against classical random initialization (RI). 

\begin{figure*}[t!]
    \centering
    \begin{subfigure}[t]{0.6\textwidth}
        \centering
        \includegraphics[width=\textwidth]{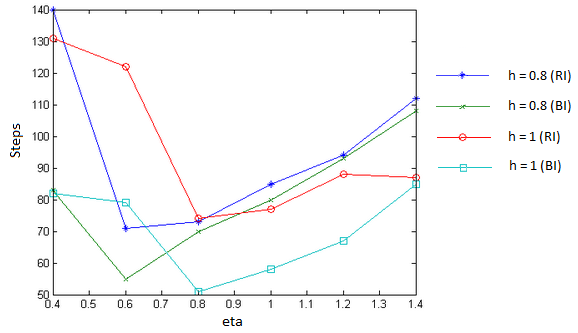}
        \caption{\label{fig:bi-ri1}}
    \end{subfigure}%
    ~ 
    \begin{subfigure}[t]{0.6\textwidth}
        \centering
        \includegraphics[width=\textwidth]{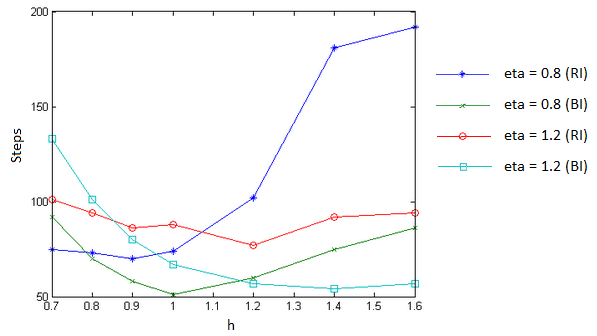}
        \caption{\label{fig:bi-ri2}}
    \end{subfigure}
    \caption{Comparison between BI and RI for convergence of BP neural net with $L=3$, $N(1)=315$, $N(2)=100$, $N(3)=85$, and hyperbolic tangent activation function}
\end{figure*}

Firstly, we use a training set composed by 26 latin printed characters for 7 different fonts (Arial, Cambria, Courier, Georgia, Tahoma, Times New Roman, Verdana), 24 greek letters, and 35 miscellaneous mathematical symbols, with 12 pt. In Figures \ref{fig:bi-ri1} and \ref{fig:bi-ri2}, performances of BP algorithm with BI and RI are compared for different values of $h$ and $\eta$, where $(-h,h)$ is the interval where weights are sampled and $\eta$ is the learning rate of the BP algorithm. The improvement in convergence rate due to BI is noticeable at a glance in these figures. In particular, we can see that BI approach is more resistant than RI with respect to high values of $h$. In fact, for large values of $h$, weights can range over a large interval. Consequently, RI produces weights scattered on a large interval causing a slower convergence of BP algorithm. On the other hand, BI seems to set initial weights on regions that allow a faster convergence of BP algorithm, despite the size of $h$. This could be very useful in complex problems where small values of $h$ do not allow convergence of BP algorithm and large intervals are necessary, as the case of the realization of an OCR for both text and formulae, where the number of different patterns for training the neural net is very high and large values of $h$ are necessary. Indeed, we can observe in our simulations that the best performances are generally obtained by BI with large values of $h$.

In Table \ref{table:mnist}, we have trained neural networks on the MNIST training set \cite{Mnist}, composed by 60000 handwritten digits. Note that in the case of the MNIST database, if training is accomplished over all the training dataset, then BP algorithm for multilayer neural networks yields a very high accuracy in the recognition on the MNIST validation set composed by 10000 handwritten digits (more than $99\%$, see, e.g., \cite{Cir}).

\begin{table}[ht]\small  
\centering
\tabcolsep=0.15cm
\caption{Comaprison between BI and RI for convergence of BP neural net trained on the MNIST database}
\scalebox{0.82}{
\begin{tabular}{|c||c|c|c|c|c|c|c|c|}
\hline 
 & \multicolumn{2}{|c|}{$L=5$, $\eta=1.5$} & \multicolumn{2}{c|}{$L=3$, $\eta=3$}\\
 \hline
 & \multicolumn{1}{|c|}{Random in.} & \multicolumn{1}{|c|}{Bayes in.} & \multicolumn{1}{|c|}{Random in.} & \multicolumn{1}{c|}{Bayes in.}\\
 \hline
h  & Steps & Steps & Steps & Steps \cr \hline
0.7 & 832 & 809 & 874 & 868 \cr \hline
0.8 & 823 & 812 & 652 & 631 \cr \hline 
0.9 & 748 & 696 & 722 & 706 \cr \hline
1 &   749 & 671 & 688 & 564 \cr \hline
1.1 & 961 & 929 & 803 & 658 \cr \hline
1.2 & 1211 & 1118 & 967 & 872 \cr \hline
\end{tabular}
}
\label{table:mnist}
\end{table}

The computational complexity to implement the classical Kalman filter is polynomial (see, e.g., \cite{Fault} p. 226). Our customization is faster since it involves fewer number of operations (matrix multiplications) than usual Kalman filter and we use circulant matrices, whose inverse can be evaluated in a very fast way. Indeed, these matrices can be diagonalized by using the Discrete Fourier Transform (\cite{Toe}, p. 32); the Discrete Fourier Transform and the inverse of a diagonal matrix are immediate to evaluate. Thus, our initialization algorithm is faster than classical Kalman filter, moreover it is iterated for a low number of steps (usually twice). Surely, this approach has a time complexity greater than random initialization. However, looking at BP Algorithm, we can observe that Eqs. \ref{eq:alg} involve similar operations (i.e., matrix multiplications or multiplications between matrices and vectors) in a minor quantity; moreover it needs a smaller number of cycles. Furthermore, we have seen that BI generally leads to a noticeable decrease of steps necessary for the convergence of the BP algorithm with respect to random initialization. Thus, using BI we can reach a faster convergence, in terms of time, of the BP algorithm than using random initialization.

Finally, we report the results on the segmentation of 296 touching characters by means of the fuzzy method explained in the previous section. The dataset is composed by 66 touching characters font Verdana and 10 pt, 58 with font Times New Roman and 20 pt, 92 font Lucida and 25 pt, 40 Georgia and 20 pt, 54 Cambria and 20 pt. Our method correctly segments the $93.6\%$ of these touching characters, improving performances obtained by using functions $g$ and $h$ one at a time that perform a correct segmentation in $76.5\%$ and $71.1\%$ of cases, respectively.

\section{Conclusion}\label{sec:conc}
We propose the development of an OCR able to recognize both alphabet characters and mathematical symbols. The pattern recognition algorithm is based on ANNs trained by means of an improvement of the backpropagation algorithm. The image segmentation algorithm is based on a fuzzy routine that combines some features usually exploited one at a time. Note that we are not proposing an OCR already completed and tuned. The proposed experimental results and simulations show that the approaches presented in this paper introduce improvements and benefits if compared with existing standard achievements.


Future works will deal with an extensive validation of the proposed system with visually impaired and blind subjects. Moreover, we will develop a novel architecture for character recognition, based on an array of ANNs which are applied in parallel to a given input pattern. This choice will enable us to reach better accuracy with respect to more traditional approaches based on a single ANN, with at the same time no appreciable degradation on performances.

\end{document}